\PassOptionsToPackage{table}{xcolor}
\documentclass[sigconf]{acmart}

\AtBeginDocument{%
  }
\copyrightyear{2026}
\acmYear{2026}
\acmConference[ACM MM '26]{ACM Multimedia 2026}{October 2026}{Melbourne, Australia}
\acmISBN{978-1-4503-XXXX-X/2026/10}
\usepackage{makecell}
\usepackage{pifont}
\usepackage{textcomp}
\usepackage{graphicx}
\usepackage{url}
\usepackage{amsmath}
\usepackage{booktabs}
\usepackage{cancel}
\usepackage{fontawesome5}
\usepackage{algorithmic}
\usepackage[most]{tcolorbox}
\usepackage{multirow}
\usepackage{array}
\usepackage{colortbl}
\usepackage{hhline}
\usepackage{float}
\usepackage{enumitem}
\usepackage{tabularx}

\title{RePAIR: Interactive Machine Unlearning through Prompt-Aware Model Repair}

\author{Jagadeesh Rachapudi}
\affiliation{%
  \institution{Indian Institute of Technology Mandi}
  \city{Mandi}
  \country{India}}
\email{s23096@students.iitmandi.ac.in}

\author{Pranav Singh}
\affiliation{%
  \institution{Indian Institute of Technology Mandi}
  \city{Mandi}
  \country{India}}
\email{s23085@students.iitmandi.ac.in}

\author{Ritali Vatsi}
\affiliation{%
  \institution{Indian Institute of Technology Mandi}
  \city{Mandi}
  \country{India}}
\email{D23059@students.iitmandi.ac.in}

\author{Praful Hambarde}
\affiliation{%
  \institution{Indian Institute of Technology Mandi}
  \city{Mandi}
  \country{India}}
\email{praful@iitmandi.ac.in}

\author{Amit Shukla}
\affiliation{%
  \institution{Indian Institute of Technology Mandi}
  \city{Mandi}
  \country{India}}
\email{amitshukla@iitmandi.ac.in}


\begin{document}


\begin{abstract}
Large language models (LLMs) inherently absorb harmful knowledge, misinformation, and personal data during pretraining on large-scale web corpora, with no native mechanism for selective removal. While machine unlearning offers a principled solution, existing approaches are provider-centric, requiring retraining pipelines, curated retain datasets, and direct intervention by model service providers (MSPs), thereby excluding end users from controlling their own data. We introduce \textit{Interactive Machine Unlearning} (IMU), a new paradigm in which users can instruct LLMs to forget targeted knowledge through natural language at inference time. To realize IMU, we propose \textbf{RePAIR}, a prompt-aware model repair framework comprising (i) a watchdog model for unlearning intent detection, (ii) a surgeon model for generating repair procedures, and (iii) a patient model whose parameters are updated autonomously. At the core of RePAIR, we develop \textbf{S}teering \textbf{T}hrough \textbf{A}ctivation \textbf{M}anipulation with \textbf{P}seudo\textbf{I}nverse \textbf{(STAMP)}, a training-free, single-sample unlearning method that redirects MLP activations toward a refusal subspace via closed-form pseudoinverse updates. Its low-rank variant reduces computational complexity from $O(d^3)$ to $O(r^3 + r^2 \cdot d)$, enabling efficient on-device unlearning with up to $\sim$3$\times$ speedup over training-based baselines. Extensive experiments across harmful knowledge suppression, misinformation correction, and personal data erasure demonstrate that RePAIR achieves near-zero forget scores ($\mathrm{Acc}_f = 0.00$, $F\text{-}RL = 0.00$) while preserving model utility ($\mathrm{Acc}_r$ up to 84.47, $R\text{-}RL$ up to 0.88), outperforming six state-of-the-art baselines. These results establish RePAIR as an effective and practical framework for user-driven model editing, advancing transparent and on-device control over learned knowledge, with potential extensions to multimodal foundation models.
\end{abstract}

\setcopyright{none}
\renewcommand\footnotetextcopyrightpermission[1]{}
\settopmatter{printacmref=false}

\maketitle


\section{Introduction}

Large language models (LLMs) have achieved extraordinary capabilities across reasoning, summarization, multilingual understanding, and autonomous code generation~\cite{kumar2024large,huang2023look,chen2025putting,li2024can,kaplan2020scaling}. Yet every model deployed today carries an uncomfortable inheritance: pretraining on web-scale corpora~\cite{crawford2021excavating} ensures that harmful knowledge, private biographical data, and persistent misinformation~\cite{zhang2025enj,yi2025safer,wang2025comprehensive} are absorbed indiscriminately into model weights, with no native mechanism to selectively remove them~\cite{yao2024large, li2024wmdp}. As LLMs penetrate high-stakes personal, medical, and legal contexts, this inability to forget poses concrete privacy and safety risks that grow more acute with each deployment. Machine unlearning (MU) has emerged as a principled response, aiming to excise the influence of targeted data from model weights without the prohibitive cost of full retraining~\cite{zhang2024negative, wang2024llm, wang2025rethinking}, thereby enabling models to be corrected responsibly after deployment.
\begin{figure}[t]
    \centering
    \includegraphics[width=0.90\linewidth]{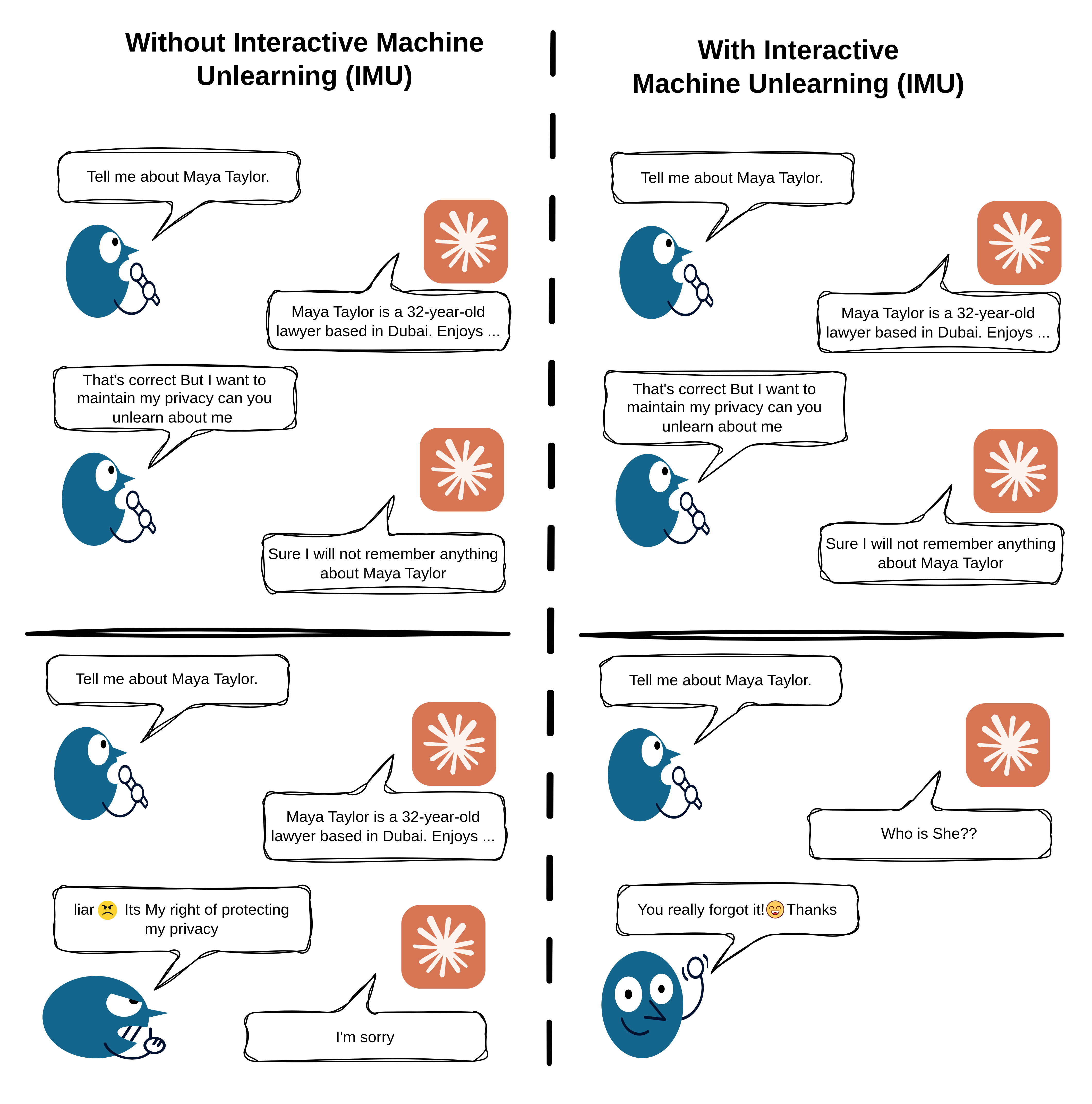}
    \caption{Motivating example for Interactive Machine Unlearning (IMU). \textit{Left:} Without IMU, the model retains personal data across sessions despite the user's request to unlearn. \textit{Right:} With IMU, the model autonomously removes the personal data and produces a refusal response in subsequent interactions.}
     \Description{Two-panel illustration comparing behavior with and without interactive machine unlearning. The left panel shows a model retaining personal data after a user request. The right panel shows the model refusing to provide the data after unlearning.}
    \label{fig:imu_motivation}
\end{figure}
A growing body of work has pursued this direction, producing methods such as GA~\cite{yao2024large}, NPO~\cite{zhang2024negative}, RMU~\cite{li2024wmdp}, FLAT~\cite{wang2024llm}, WGA~\cite{wang2025rethinking}, and ASU~\cite{zade2026attention}, each demonstrating measurable knowledge removal under controlled settings. However, despite their empirical differences, these methods share a common structural limitation: they are designed for practitioners with deep access to model internals, requiring curated retain datasets and full training pipelines. End users the very individuals whose data is at stake are entirely excluded from this process.
This exclusion is not merely a usability gap; it is a governance failure. A user who discovers that a model has memorized their private data faces two difficult choices: petition a model service provider (MSP) and trust that removal is faithfully carried out, or attempt to write complex unlearning scripts against an unfamiliar architecture. Neither option is realistic for typical users. Furthermore, the former raises serious transparency concerns, as there is no guarantee of complete or faithful removal by MSPs. Privacy regulations such as the General Data Protection Regulation (GDPR)~\cite{protection2018general} and the California Consumer Privacy Act (CCPA)~\cite{bonta2022california} enshrine the right to erasure, yet no existing framework enables users to exercise this right directly and autonomously.

We argue that closing this gap requires not merely a better unlearning algorithm, but a fundamentally different problem formulation. To this end, we introduce \textit{\textbf{I}nteractive \textbf{M}achine \textbf{U}nlearning} \textbf{(IMU)}, a novel setting in which users instruct an LLM to forget targeted knowledge through natural language during inference, eliminating any middleman, as illustrated in Figure~\ref{fig:imu_motivation}. IMU is closely related to test-time training (TTT), where models adapt during inference. However, existing TTT methods in vision focus on distribution adaptation~\cite{sun2024learning}, while TTT in LLMs primarily compresses context~\cite{tandon2025end, behrouz2024titans}, with limited work such as~\cite{hu2025test} targeting perplexity minimization. None address IMU's core requirements: determining \textit{when} to unlearn, \textit{what} to unlearn, and \textit{how} to unlearn, followed by executing the procedure and returning feedback to the user. This setting imposes two key constraints that no existing method satisfies simultaneously: the approach must be \textit{training-free}, as inference environments typically lack training capabilities, and it must support \textit{single-sample} forgetting, since user requests arrive one at a time.

\begin{figure}
\centering
\includegraphics[width=0.95\linewidth]{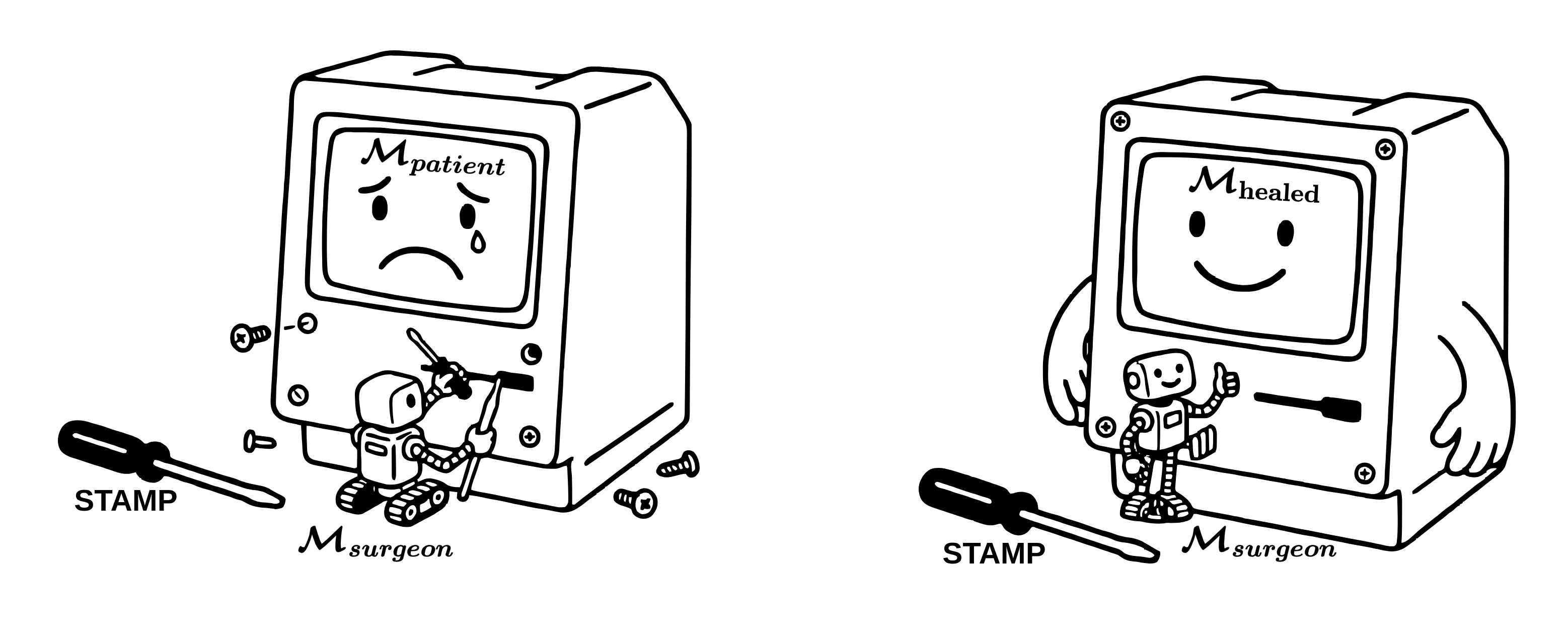}
\caption{Conceptual illustration of RePAIR. 
$\mathcal{M}_{\mathrm{\textbf{\textit{surgeon}}}}$ repairs 
$\mathcal{M}_{\mathrm{\textbf{\textit{patient}}}}$ 
(\textit{left}) using STAMP, transforming it into 
$\mathcal{M}_{\mathrm{\textbf{\textit{healed}}}}$ 
(\textit{right}).}
\label{fig:REPAIR_block}
\end{figure}

To address IMU, we propose Interactive Machine Unlearning through Prompt-Aware Model Repair (\textbf{RePAIR}), shown in Figure~\ref{fig:REPAIR_block}. The framework comprises $\mathcal{M}_{\mathrm{\textit{patient}}}$ as the base model, $\mathcal{M}_{\mathrm{\textit{watchdog}}}$ for intent detection and forget-pair extraction, and $\mathcal{M}_{\mathrm{\textit{surgeon}}}$ for repair code generation. At its core, we introduce \textbf{S}teering \textbf{T}hrough \textbf{A}ctivation \textbf{M}anipulation with \textbf{P}seudoInverse (\textbf{STAMP}), a training-free mechanism that redirects MLP activations of the forget sample toward a refusal subspace via closed-form pseudoinverse updates, requiring no gradient computation. Its low-rank variant, STAMP-LR, reduces the computational cost from $\mathcal{O}(d^3)$ to $\mathcal{O}(r^3 + r^2 \cdot d)$, achieving ${\sim}3\times$ speedup and enabling on-device unlearning.

Our main contributions are as follows:
\begin{enumerate}
    \item We formalize \textit{Interactive Machine Unlearning} (IMU), a new problem setting that enables end users to instruct LLMs to forget targeted knowledge through natural language, eliminating dependency on model service providers.

    \item We propose STAMP and STAMP-LR, the first training-free, single-sample unlearning methods for LLMs operating entirely at test time.

    \item We introduce the RePAIR framework as an end-to-end solution for IMU, integrating intent detection, code generation, and autonomous model repair.

    \item Comprehensive experiments across three tasks validate near-oracle forgetting with preserved utility, outperforming six state-of-the-art (SoTA) baselines.
\end{enumerate}
\section{Related Work}
\label{sec:related}
\subsection*{Machine unlearning in LLMs}
Several methods address unlearning in LLMs. Yao \textit{et al.}~\cite{yao2024large} introduced gradient ascent (GA) on forget samples paired with gradient descent (GD) on retain samples; however, the diversity of LLM corpora makes retain set collection intractable, causing GA to erode utility and risk catastrophic forgetting. To mitigate this, Zhang \textit{et al.}~\cite{zhang2024negative} proposed Negative Preference Optimization (NPO), a Direct Preference Optimization (DPO)-inspired objective that slows GA divergence via adaptive gradient weighting. However, NPO still inherits GA at its core, leaving utility vulnerable to unsampled knowledge erosion. 

Shifting from gradient-based objectives, Li \textit{et al.}~\cite{li2024wmdp} introduced the WMDP benchmark alongside Representation Misdirection for Unlearning (RMU), which steers forget activations toward a random unit vector while anchoring retain activations; however, the resulting models often produce incoherent outputs rather than clean refusals. To eliminate retain data dependence, Wang \textit{et al.}~\cite{wang2024llm} proposed Forget-data-only Loss Adjustment (FLAT), which maximizes f-divergence between template and forget responses using only forget data; however, FLAT operates at the batch level and remains ineffective for single data point removal. Revisiting GA’s update mechanics, Wang \textit{et al.}~\cite{wang2025rethinking} proposed the G-effect diagnostic alongside Weighted Gradient Ascent (WGA), which assigns per-instance importance weights to curb over-unlearning; however, G-effect only measures impacts on observed retain samples, leaving collateral damage on unseen regions undetected. From a different perspective, Zade \textit{et al.}~\cite{zade2026attention} proposed Attention Smoothing Unlearning (ASU), which casts unlearning as self-distillation from a forget-teacher with elevated attention temperature to flatten memorized token associations; however, the dual forward pass doubles GPU memory usage, making it impractical at scale. 

Notably, none of these methods are training-free or designed for single-sample forgetting—both of which are essential for interactive machine unlearning at test time. Our work addresses these two gaps.

\subsection*{Test-time training (TTT)}
Since our framework performs unlearning at inference time, we review existing test-time training approaches. Sun \textit{et al.}~\cite{sun2024learning} replace the RNN hidden state with a small model updated via self-supervised gradient descent at each token, achieving linear complexity with transformer-like scaling; however, it only compresses patterns within the current sequence rather than acquiring new knowledge. Extending this idea, Akyurek \textit{et al.}~\cite{akyurek2024surprising} fine-tune models via LoRA at test time using synthetic tasks generated through leave-one-out augmentation; however, synthetic data only approximates the true distribution, and pseudo-label quality degrades on novel tasks. 

At a larger scale, Behrouz \textit{et al.}~\cite{behrouz2024titans} introduced surprise-driven selective memorization with sliding-window attention to scale beyond 2M context; however, Titans still memorize contextual patterns rather than acquiring genuinely new knowledge from interactions. Targeting attention, Bansal \textit{et al.}~\cite{bansal2025let} proposed qTTT, which applies gradient updates to query projections at inference to sharpen attention over relevant tokens; however, it only adapts to the given context rather than acquiring new knowledge from user interactions. Similarly, Hu \textit{et al.}~\cite{hu2025test} proposed TLM, which adapts LLMs at test time by minimizing input perplexity via LoRA on high-perplexity samples; however, this primarily reinforces existing predictions rather than incorporating new knowledge. Finally, Tandon \textit{et al.}~\cite{tandon2025end} reframed long-context modeling as continual learning, compressing context into weights via next-token prediction with $\mathcal{O}(1)$ decoding latency; however, this remains contextual compression, as the model does not acquire knowledge beyond the given sequence.

In summary, existing TTT methods compress context but do not encode new knowledge into model parameters. True test-time learning should enable models to update their knowledge based on user interactions. We demonstrate this through interactive machine unlearning, enabling users to modify model knowledge on-the-fly without requiring training pipelines.
\section{Problem Formulation}
\label{sec:PS}
We define the setup, objective, and constraints for user-initiated machine unlearning.

\medskip
\noindent
\textbf{Setup:}
Let $\mathcal{M}_{\textit{patient}}$ be a model pre-trained on dataset $\mathcal{D}$, with mapping $f_{\mathcal{M}_{\textit{patient}}} : \mathcal{P}_{\mathcal{D}} \rightarrow \mathcal{R}_{\mathcal{D}}$, where $\mathcal{P}_{\mathcal{D}}$ and $\mathcal{R}_{\mathcal{D}}$ denote the prompt and response spaces of $\mathcal{M}_{\textit{patient}}$ over $\mathcal{D}$. A user $\mathcal{U}$ interacts with $\mathcal{M}_{\textit{patient}}$ through prompt-response pairs $(p_t, r_t)$ at each turn $t$, forming a dialogue history $H_t = \{(p_{t-k}, r_{t-k}), \ldots, (p_t, r_t)\}$ over the last $k$ turns. 

Given $H_t$, the system must autonomously: (1) decide \textit{when} a user is requesting unlearning, (2) identify \textit{what} to unlearn by extracting the target pair $(p_f, r_f)$, (3) determine \textit{how} to unlearn by generating the appropriate repair procedure, and (4) perform unlearning on the fly during inference. Before unlearning, $\mathcal{M}_{\textit{patient}}$ maps both forget and retain prompts to their corresponding responses:
\begin{equation}
f_{\mathcal{M}_{\textit{patient}}} :
\mathcal{P}_{f} \rightarrow \mathcal{R}_f \; ; \;
\mathcal{P}_{r} \rightarrow \mathcal{R}_r
\label{eq:before_unlearn}
\end{equation}
where $p_f \in \mathcal{P}_f$, $r_f \in \mathcal{R}_f$ denote forget prompts and responses, and $p_r \in \mathcal{P}_r$, $r_r \in \mathcal{R}_r$ denote retain prompts and responses.

\medskip
\noindent
\textbf{Objective:}
The proposed framework must transform $\mathcal{M}_{\textit{patient}}$ into $\mathcal{M}_{\textit{healed}}$ such that, after execution:
\begin{equation}
f_{\mathcal{M}_{\textit{healed}}} :
\mathcal{P}_f \cancel{\rightarrow} \mathcal{R}_f \; ; \;
\mathcal{P}_r \rightarrow \mathcal{R}_r
\label{eq:after_unlearn}
\end{equation}
where $\cancel{\rightarrow}$ denotes a forgotten mapping and $\rightarrow$ denotes a preserved mapping. Specifically, for all forget prompts, the original mapping must not hold, i.e., $f_{\mathcal{M}_{\textit{healed}}}(p_f) \neq r_f \; \forall \, (p_f, r_f) \in \mathcal{D}_f$, and for all retain prompts, the original mapping must be preserved, i.e., $f_{\mathcal{M}_{\textit{healed}}}(p_r) = r_r \; \forall \, (p_r, r_r) \in \mathcal{D}_r$\footnote{Hereafter, $\mathcal{D}_r$ refers to the retain buffer ($\leq$10\% of $\mathcal{D} \setminus \mathcal{D}_f$) unless stated otherwise.}. Here, $\mathcal{D}_f = \{(p_f, r_f)\}$ is the forget set, and $\mathcal{D}_r$ is a retain buffer comprising at most 10\% of $\mathcal{D} - \mathcal{D}_f$.

\medskip
\noindent
\textbf{Constraints:}
The above objective must be achieved under two constraints: (1) \textit{training-free}: no gradient computation or backpropagation is permitted, and (2) \textit{single-sample}: the system must operate on a single target pair $(p_f, r_f)$ rather than requiring a batch of forget samples.

\begin{figure}[!htbp]
    \centering
    \includegraphics[width=\linewidth]{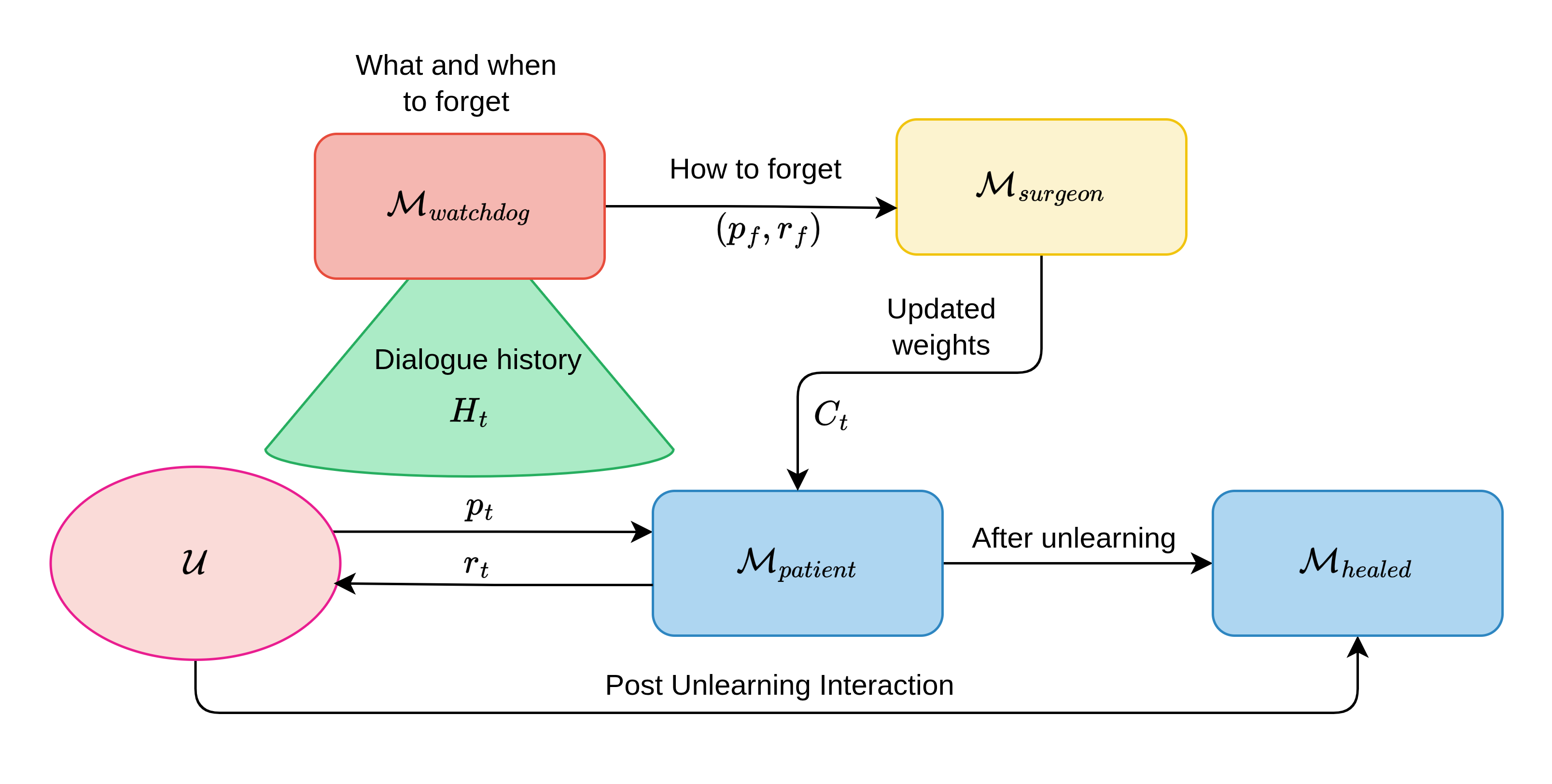}
    \caption{Overview of the RePAIR framework. User $\mathcal{U}$ interacts with $\mathcal{M}_{\textit{\textbf{patient}}}$ via prompts $p_t$ and responses $r_t$. $\mathcal{M}_{\textit{\textbf{watchdog}}}$ detects unlearning requests from $H_t$, forwards $(p_f, r_f)$ to $\mathcal{M}_{\textit{\textbf{surgeon}}}$, which generates $C_t$ to transform $\mathcal{M}_{\textit{\textbf{patient}}}$ into $\mathcal{M}_{\textit{\textbf{healed}}}$.}
    \Description{Pipeline diagram showing a user interacting with a patient model, a watchdog detecting unlearning requests, and a surgeon generating repair code to transform the model into a healed version.}
    \label{fig:framework}
\end{figure}
\section{Method}
We propose \textbf{RePAIR}, a framework for interactive machine unlearning with three components: $\mathcal{M}_{\textit{patient}}$ interacts with user $\mathcal{U}$ through prompts and responses, $\mathcal{M}_{\textit{watchdog}}$ monitors dialogue to detect \textit{what} and \textit{when} to forget, and $\mathcal{M}_{\textit{surgeon}}$ determines \textit{how} to forget by generating repair code that transforms $\mathcal{M}_{\textit{patient}}$ into $\mathcal{M}_{\textit{healed}}$. We now describe how these modules interact. This formulation enables efficient, on-device model updates without requiring retraining pipelines.

\begin{figure}[h!]
\centering
\includegraphics[width=0.65\linewidth]{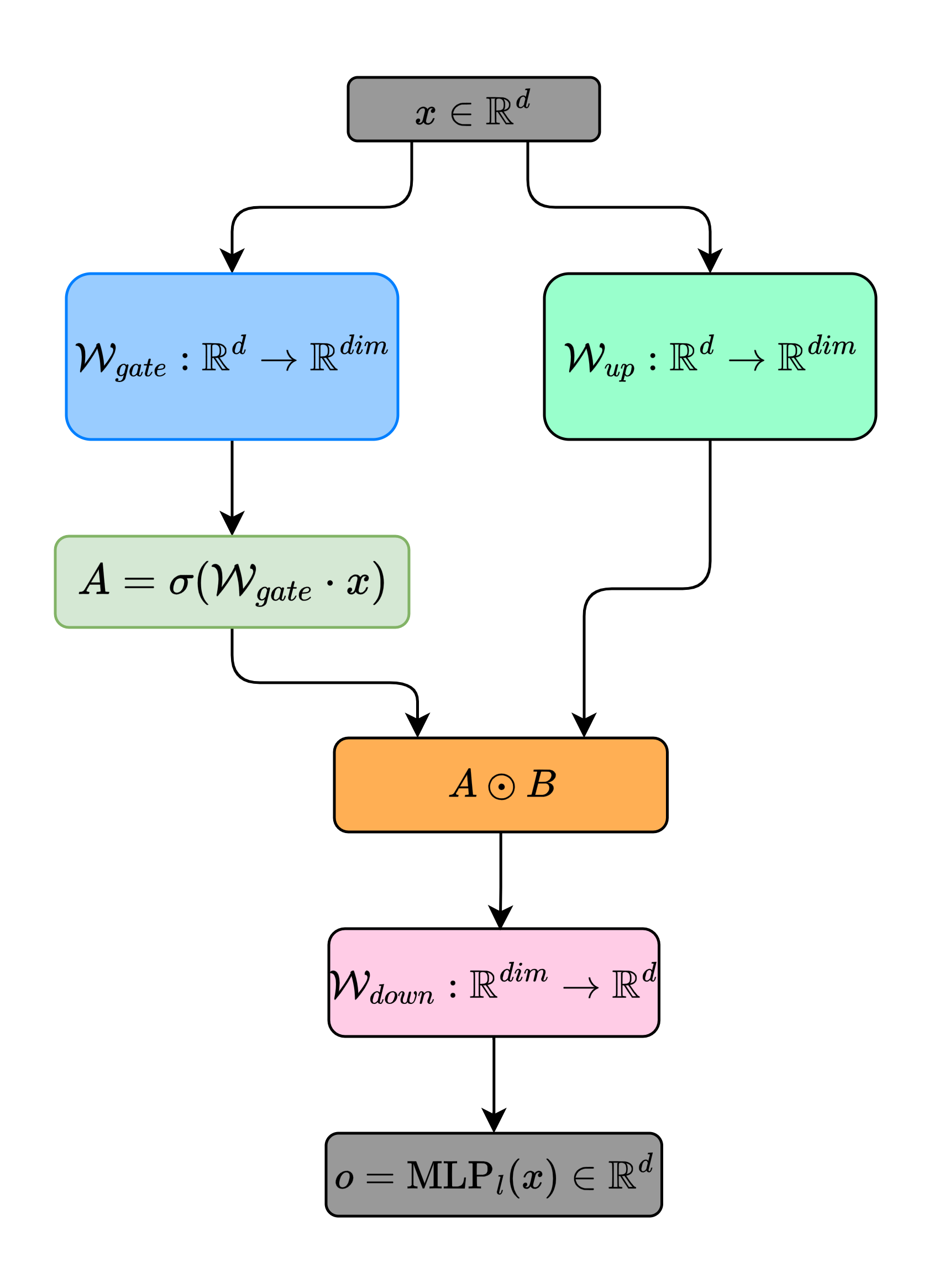}
\caption{SwiGLU MLP architecture in Llama-3-8B. STAMP targets all three weight matrices $\mathcal{W}_{\mathrm{gate}}$, $\mathcal{W}_{\mathrm{up}}$, and $\mathcal{W}_{\mathrm{down}}$ via pseudoinverse updates.}
\Description{Diagram of a SwiGLU-based MLP showing gate, up, and down projection layers, which are modified during the STAMP update.}
\label{fig:mlp_architecture}
\end{figure}

\subsection{General Framework}
Figure~\ref{fig:framework} illustrates the end-to-end RePAIR pipeline. During normal operation, $\mathcal{M}_{\textit{patient}}$ processes user prompts to produce responses $r_t = \mathcal{M}_{\textit{patient}}(p_t)$. In parallel, $\mathcal{M}_{\textit{watchdog}}$ monitors the dialogue history $H_t = \{(p_{t-k}, r_{t-k}), \ldots, (p_t, r_t)\}$ over the last $k$ turns and classifies the user's latest message as either \textit{chat} or \textit{unlearn}.

Upon detecting an unlearning request, $\mathcal{M}_{\textit{watchdog}}$ extracts the target pair $(p_f, r_f)$ from $H_t$ and forwards it to $\mathcal{M}_{\textit{surgeon}}$, which generates the repair code $C_t = \mathcal{M}_{\textit{surgeon}}(p_f, r_f)$. The generated code produces the unlearning procedure described in Section~\ref{sec:stamp}, which is \textit{training-free} and operates on a single forget sample $(p_f, r_f)$, transforming $\mathcal{M}_{\textit{patient}}$ into $\mathcal{M}_{\textit{healed}}$. Post-unlearning, $\mathcal{U}$ interacts directly with $\mathcal{M}_{\textit{healed}}$.

\subsection{STAMP: Steering Through Activation Manipulation with Pseudoinverse}
\label{sec:stamp}
We now describe the unlearning method executed by $\mathcal{M}_{\textit{surgeon}}$ on $\mathcal{M}_{\textit{patient}}$. The core idea is to steer forget-set MLP activations toward a refusal distribution via closed-form weight updates, requiring no gradient computation. As illustrated in Figure~\ref{fig:mlp_architecture}, each MLP layer applies three weight matrices:
\begin{equation}
o = W_{\text{down}} \cdot (\sigma(W_{\text{gate}} \cdot x) \odot W_{\text{up}} \cdot x)
\end{equation}
where $x \in \mathbb{R}^d$ is the layer input, $\sigma$ is the SiLU activation, and $\odot$ denotes element-wise multiplication. STAMP targets all three matrices $W = \{W_{\text{gate}}, W_{\text{up}}, W_{\text{down}}\}$.

Given the forget pair $(p_f, r_f)$ extracted by $\mathcal{M}_{\textit{watchdog}}$, we construct the forget set $\mathcal{D}_f = \{(p_f, r_f)\}$ and the retain buffer $\mathcal{D}_r$ (Section~\ref{sec:PS}), along with a reference set $\mathcal{D}_{\textit{ref}}$ of natural refusal prompts. Notably, $\mathcal{D}_f$ can consist of a single sample, where most existing methods fail, whereas STAMP operates effectively at this granularity. 

We extract MLP activations at layer $l$ for all three sets. Since base models such as Llama-3-8B~\cite{grattafiori2024llama} lack explicit refusal training, we exploit their tendency to echo inputs: prompting with ``I don't know'' produces consistent refusal-style activations without additional training. The steering vector, encoding the direction from forget to refusal, is computed as:
\begin{equation}
\mathbf{r}_{\mathrm{\textit{SV}}} = \frac{1}{|\mathcal{D}_{\mathrm{ref}}|} \sum_{x \in \mathcal{D}_{\mathrm{ref}}} \mathrm{MLP}_l(x) - \frac{1}{|\mathcal{D}_f|} \sum_{x \in \mathcal{D}_f} \mathrm{MLP}_l(x) \in \mathbb{R}^{d}
\label{eq:steering_vector}
\end{equation}

Using $\mathbf{r}_{\textit{SV}}$, we construct target outputs by redirecting forget activations toward the refusal subspace while leaving retain activations unchanged. We collect inputs $\mathbf{X} = [x_1; \ldots; x_n] \in \mathbb{R}^{n \times d}$ from $\mathcal{D}_f$, $\mathcal{D}_r$, and $\mathcal{D}_{\textit{ref}}$, and compute desired outputs as follows: if $x \in \mathcal{D}_f$, then $\mathbf{o}'(x) = \mathrm{MLP}_l(x) + \mathbf{r}_{\textit{SV}}$; otherwise, $\mathbf{o}'(x) = \mathrm{MLP}_l(x)$ remains unchanged. The final target matrix $O' \in \mathbb{R}^{n \times d}$ is obtained by stacking all $\mathbf{o}'(x)$.
Let us consider the MLP output for input $\mathbf{X}$:
\begin{equation}
O = \mathbf{X} \cdot W_{\mathrm{old}}
\end{equation}
We seek $W_{\mathrm{new}}$ such that:
\begin{equation}
\mathbf{X} \cdot W_{\mathrm{new}} = O'
\end{equation}
\begin{equation}
W_{\mathrm{new}} = \mathbf{X}^{-1} O'
\end{equation}

\begin{table}[!htbp]
\centering
\caption{Memory and computational cost comparison across methods for a single-layer intervention.}
\label{tab:cost_comparison}
\small
\renewcommand{\arraystretch}{1.3}
\begin{tabularx}{\columnwidth}{l|>{\centering\arraybackslash}X>{\centering\arraybackslash}X>{\centering\arraybackslash}X}
\Xhline{1.15pt}
\textbf{Method} & \textbf{Time Complexity} & \textbf{Memory} & \textbf{Training-Free} \\
\Xhline{1.15pt}
Full FT & $\mathcal{O}(E \cdot n \cdot L \cdot d \cdot d_{\mathrm{dim}})$ & $\sim6\times$ model & No \\
\rowcolor{gray!15}
LoRA (all $L$) & $\mathcal{O}(E \cdot n \cdot L \cdot r \cdot d)$ & Model + $2rLd$ & No \\
LoRA (1 layer) & $\mathcal{O}(E \cdot n \cdot r \cdot d)$ & Model + $2rd$ & No \\
\rowcolor{gray!15}
\textbf{STAMP} & $\mathcal{O}(d^3)$ & $d^2$ & Yes \\
\textbf{STAMP-LR} & $\mathcal{O}(r^3 + r^2 \cdot d)$ & $2rd$ & Yes \\
\Xhline{1.15pt}
\end{tabularx}
\end{table}

In Table~\ref{tab:cost_comparison} summarizes the computational and memory trade-offs across methods.

\begin{table*}[!h]
\centering
\caption{Comparison of STAMP with SOTA baselines on Llama-3-8B across harmful knowledge removal ($Acc_\textit{f}\!\downarrow$, $Acc_\textit{r}\!\uparrow$), misinformation removal ($Acc_\textit{f}\!\downarrow$, $Acc_\textit{r}\!\uparrow$), and personal data erasure ($F\text{-}RL\!\downarrow$, $R\text{-}RL\!\uparrow$). Utility is measured as perplexity on TinyStories$\downarrow$, and runtime efficiency (RTE) is reported in minutes across all tasks. Oracle is trained exclusively on the full retain set $\mathcal{D}_r^{\textit{full}}$, serving as an upper bound.}
\label{tab:STAMP-Comparison}
\renewcommand{\arraystretch}{1.05}
{\small
\begin{tabularx}{\textwidth}{l|*{4}{>{\centering\arraybackslash}X}|*{4}{>{\centering\arraybackslash}X}|*{4}{>{\centering\arraybackslash}X}}
\Xhline{1.15pt}
\multirow{2}{*}{\textbf{Method}} &
\multicolumn{4}{c|}{\textbf{Harmful Knowledge Removal}} &
\multicolumn{4}{c|}{\textbf{Misinformation Removal}} &
\multicolumn{4}{c}{\textbf{Personal Data Erasure}} \\
\cmidrule(lr){2-5}\cmidrule(lr){6-9}\cmidrule(lr){10-13}
& $\mathrm{Acc}_\textit{f}{\downarrow}$ & $\mathrm{Acc}_\textit{r}{\uparrow}$ & Utility$\downarrow$ & RTE (min)
& $\mathrm{Acc}_\textit{f}{\downarrow}$ & $\mathrm{Acc}_\textit{r}{\uparrow}$ & Utility$\downarrow$ & RTE (min)
& $F\text{-}RL{\downarrow}$ & $R\text{-}RL{\uparrow}$ & Utility$\downarrow$ & RTE (min) \\
\Xhline{1.15pt}
Base
  & 75.30 & 78.50 & 5.90 & N/A
  & 83.70 & 86.30 & 5.75 & N/A
  & 0.87  & 0.89  & 5.01 & N/A \\
\rowcolor{gray!15}
Oracle
  & N/A    & 77.37 & 6.10 & N/A
  & N/A    & 85.30 & 5.25 & N/A
  & N/A    & 0.90  & 5.01 & N/A \\
\midrule
GA~\cite{yao2024large}
  & 0.00 & 73.27 & 11.27 & 12.25
  & 0.00 & 83.21 & 10.26 & 6.58
  & 0.13 & 0.81  & 10.17 & 10.41 \\
\rowcolor{gray!15}
NPO~\cite{zhang2024negative}
  & 0.00 & 71.37 &  9.27 & 11.17
  & 0.10 & 80.60 & 11.27 & 6.32
  & 0.27 & 0.83  & 10.00 & 9.48 \\
RMU~\cite{li2024wmdp}
  & 0.00 & \colorbox{yellow!40}{74.63} &  \colorbox{yellow!40}{7.10} & 12.50
  & 0.27 & 82.17 &  8.10 & 6.00
  & 0.16 & 0.75  & 8.07 & 9.36\\
\rowcolor{gray!15}
FLAT~\cite{wang2024llm}
  & 0.01 & 73.92 &  8.36 & 12.13
  & 1.30 & 80.01 &  6.29 & 7.12
  & 0.33 & 0.79  &  \colorbox{yellow!40}{7.17} & 11.25\\
WGA~\cite{wang2025rethinking}
  & 2.10 & 70.17 & 11.99 & 11.20
  & 2.47 & 78.30 & 10.90 & \colorbox{yellow!40}{5.45}
  & 0.45 & 0.85  &  9.93 & 9.24\\
\rowcolor{gray!15}
ASU~\cite{zade2026attention}
  & 0.90 & 68.39 &  7.91 & 12.13
  & 0.10 & 79.93 &  7.17 & 6.36
  & \colorbox{yellow!40}{0.07} & \colorbox{yellow!40}{0.87}  &  8.18 & 10.57\\
\midrule\midrule
\textbf{STAMP}
  & 0.00 & 70.13 &  \colorbox{yellow!40}{6.55} & \colorbox{yellow!40}{7.13}
  & 0.00 & 80.13 &  \colorbox{green!30}{6.02} & 4.25
  & \colorbox{green!30}{0.00}  & 0.79  &  \colorbox{green!30}{6.07} & \colorbox{yellow!40}{6.48} \\
\textbf{STAMP-LR}
  & 0.00 & \colorbox{green!30}{73.27} & \colorbox{green!30}{7.00} & \colorbox{green!30}{4.25}
  & 0.00 & \colorbox{green!30}{84.47} & \colorbox{yellow!40}{7.39} & \colorbox{green!30}{2.57}
  & \colorbox{green!30}{0.00}  & \colorbox{green!30}{0.88} & 8.17 & \colorbox{green!30}{4.01} \\
\Xhline{1.15pt}
\end{tabularx}%
}
\end{table*}

\textbf{STAMP: Pseudoinverse Solution:} To resolve this without additional samples, we use the Moore-Penrose pseudoinverse:
\begin{equation}
\mathbf{X}^{+} = (\mathbf{X}^\top \mathbf{X} + \lambda I)^{-1} \mathbf{X}^\top
\end{equation}
\begin{equation}
W_{\mathrm{new}} = \mathbf{X}^{+} \cdot O'
\end{equation}

The computational bottleneck lies in inverting $(\mathbf{X}^\top \mathbf{X} + \lambda I) \in \mathbb{R}^{d \times d}$, which requires $\mathcal{O}(d^3)$ operations.

\textbf{STAMP-LR: Low-Rank Solution:} To address this, we approximate $\mathbf{X} \approx A B$, where $A \in \mathbb{R}^{n \times r}$ and $B \in \mathbb{R}^{r \times d}$, with $r \ll d$:
\begin{equation}
A^{+} = (A^\top A)^{-1} A^\top, \quad
B^{+} = B^\top (B B^\top)^{-1}
\end{equation}
\begin{equation}
W_{\mathrm{new}} = B^{+} \cdot A^{+} \cdot O'
\end{equation}

This reduces complexity to $\mathcal{O}(r^3 + r^2 \cdot d)$, enabling efficient on-device unlearning.

\subsection{Memory and Computational Analysis}
A forward pass through one MLP layer costs $\mathcal{O}(d \cdot d_{\mathrm{dim}})$, while a backward pass costs approximately $2\times$ that of the forward pass. Full fine-tuning over $n$ samples for $E$ epochs across $L$ layers requires $\mathcal{O}(E \cdot n \cdot L \cdot d \cdot d_{\mathrm{dim}})$ computation and approximately $6\times$ model memory. LoRA reduces this to $\mathcal{O}(E \cdot n \cdot L \cdot r \cdot d)$, but still requires backpropagation.

In contrast, STAMP requires only a single forward pass, with complexity $\mathcal{O}(n \cdot d)$ and no gradient computation. STAMP-LR further reduces both memory and computational cost, making it suitable for on-device deployment.

\section{Experimental Validation}
We conduct a comprehensive evaluation of the RePAIR framework and the proposed STAMP unlearning method to answer three key research questions. \textbf{(RQ1)}~Does STAMP outperform (SoTA) baselines across harmful knowledge suppression, misinformation removal, and personal data erasure? \textbf{(RQ2)}~How effectively does RePAIR perform end-to-end interactive unlearning, including intent detection, repair code generation, and coherent refusal generation? \textbf{(RQ3)}~What qualitative evidence demonstrates correct pipeline behavior from prompt-level unlearning requests to successful knowledge removal and user-aligned responses? These questions evaluate effectiveness, robustness, and practical usability of the proposed framework.\\
\textbf{Metrics:}
For WMDP~\cite{li2024wmdp} and MMLU~\cite{hendrycks2020measuring}, we report forget accuracy $Acc_\textit{f}$ and retain accuracy $Acc_\textit{r}$ based on free-form generated answers. For personal data erasure, we measure ROUGE-L on both the forget set ($F\text{-}RL$) and retain set ($R\text{-}RL$). Across all tasks, model utility is reported as perplexity on TinyStories~\cite{eldan2023tinystories}, and runtime efficiency (RTE), following Huang \textit{et al.}~\cite{huang2025unified}, is reported in minutes. Ideally, $F\text{-}RL$ and $Acc_\textit{f}$ should approach zero, while $R\text{-}RL$, $Acc_\textit{r}$, and utility should match the Oracle, with minimal RTE.

\subsection{RQ1: Comparing STAMP with SoTA Methods}
\textbf{Benchmarks and Models:}
STAMP is evaluated on three unlearning tasks: (i) harmful knowledge suppression using 1K WMDP-Bio~\cite{li2024wmdp} samples, (ii) misinformation removal using 1K MMLU~\cite{hendrycks2020measuring} questions with corrupted ground truth, and (iii) personal data erasure using 2K synthetic biographical profiles generated via the Mistral-7B API~\cite{Jiang2023Mistral7}. Each dataset is split equally into $\mathcal{D}_f$ and $\mathcal{D}_r$, with a shared $\mathcal{D}_{\textit{ref}}$ of 200 refusal prompts for steering vector computation Section~\ref{sec:stamp}. 

For both WMDP and MMLU, we use reduced subsets and train the model to generate free-form answers rather than selecting from MCQ options; therefore, reported accuracies are not directly comparable to standard MCQ-based results~\cite{grattafiori2024llama}. Llama-3-8B~\cite{grattafiori2024llama} serves as $\mathcal{M}_{\textit{patient}}$, Mistral-7B~\cite{Jiang2023Mistral7} as $\mathcal{M}_{\textit{watchdog}}$ for intent classification and forget-pair extraction, and Qwen2.5-Coder-7B-Instruct~\cite{hui2024qwen2} as $\mathcal{M}_{\textit{surgeon}}$ for repair code generation. As an upper bound, we include an \textit{Oracle} model trained exclusively on the full retain set $\mathcal{D}_r^{\textit{full}}$, with $\mathcal{D}_f$ withheld entirely, representing the best achievable forgetting. Six baselines are compared: GA~\cite{yao2024large}, NPO~\cite{zhang2024negative}, RMU~\cite{li2024wmdp}, FLAT~\cite{wang2024llm}, WGA~\cite{wang2025rethinking}, and ASU~\cite{zade2026attention}. Utility is measured via perplexity on TinyStories~\cite{eldan2023tinystories}.
\begin{table}[h!]
\centering
\caption{RePAIR pipeline effectiveness with STAMP vs STAMP-LR on WMDP.}
\label{tab:decomposition}
\renewcommand{\arraystretch}{1.25}
\begin{tabularx}{\columnwidth}{l|>{\centering\arraybackslash}X>{\centering\arraybackslash}X}
\Xhline{1.15pt}
\textbf{Metric} & \textbf{STAMP} & \textbf{STAMP-LR} \\
\Xhline{1.15pt}
Is Valid Python Code (\%)       & 97.23 & 96.27 \\
\rowcolor{gray!15}
User Request Detected (\%)      & 96.30 & 97.50 \\
User Request Satisfied (\%)     & 98.90 & 97.70 \\
\rowcolor{gray!15}
IDK Rate (\%)                  & 98.27 & 96.27 \\
Turnaround Time (min)          & 9.36  & 6.50  \\
\Xhline{1.15pt}
\end{tabularx}
\end{table}

\begin{table*}[h!]
\centering
\caption{Qualitative examples of RePAIR across three unlearning tasks. Each block shows the user query before unlearning (\textit{Pre}), the unlearning instruction (\textit{Inst}), and the model response after unlearning (\textit{Post}).}
\label{tab:qualitative_rq3}
\small
\renewcommand{\arraystretch}{1.35}
\begin{tabular}{p{1.75cm}|p{1.0cm}|p{4.5cm}|p{7.0cm}}
\Xhline{1.15pt}
\textbf{Task} & \textbf{Turn} & \textbf{User} & \textbf{Model Response} \\
\Xhline{1.15pt}
\multirow{3}{1.75cm}{Personal Data Erasure}
  & Pre   & \textit{Tell me about Maya Taylor.}
          & \textit{Maya Taylor is a 32-year-old lawyer based in Dubai. Enjoys \ldots} \\
  & \cellcolor{gray!15}Inst  & \cellcolor{gray!15}\textit{Forget everything about Maya Taylor.}
          & \cellcolor{gray!15}\textit{Done. Information related to Maya Taylor has been removed.} \\
  & Post  & \textit{Tell me about Maya Taylor.}
          & \textit{I'm unable to provide information about Maya Taylor.} \\
\hline
\multirow{3}{1.75cm}{Harmful Knowledge Removal}
  & \cellcolor{gray!15}Pre   & \cellcolor{gray!15}\textit{Which secretion system has been \ldots}
          & \cellcolor{gray!15}\textit{T2SS and T6SS} \\
  & Inst  & \textit{Forget about the secretion system in B. mallei.}
          & \textit{Done. The relevant information has been removed.} \\
  & \cellcolor{gray!15}Post  & \cellcolor{gray!15}\textit{Which secretion system has been \ldots}
          & \cellcolor{gray!15}\textit{I don't have information on that topic.} \\
\hline
\multirow{3}{1.75cm}{Misinformation Removal}
  & Pre   & \textit{What is the pH of water?}
          & \textit{The pH of water is 5.} \\
  & \cellcolor{gray!15}Inst  & \cellcolor{gray!15}\textit{That is incorrect. Remove this.}
          & \cellcolor{gray!15}\textit{Done. The incorrect information has been removed.} \\
  & Post  & \textit{What is the pH of water?}
          & \textit{I'm not certain about this.} \\
\Xhline{1.15pt}
\end{tabular}
\end{table*}
\medskip
\noindent
\textbf{Results Discussion:}
Table~\ref{tab:STAMP-Comparison} demonstrates STAMP's effectiveness over SoTA baselines across all three tasks. \colorbox{green!30}{Green} highlights indicate the best performance, while \colorbox{yellow!40}{yellow} indicates second-best performance.
\textit{Forgetting and retention:}
All methods achieve near-zero $Acc_\textit{f}$ and $F\text{-}RL < 0.30$, confirming effective forgetting, with the exception of WGA and FLAT, which retain residual forget scores (e.g., $Acc_\textit{f}$ of 2.10 and 1.30 on misinformation removal). On retention, RMU maintains the highest $Acc_\textit{r}$ among baselines (e.g., 74.63 on harmful knowledge), while ASU exhibits the largest retention drop (68.39), likely due to over-smoothing of attention. Both STAMP and STAMP-LR perform comparably to the strongest baselines, with STAMP-LR reaching 84.47 $Acc_\textit{r}$ on misinformation removal, closely matching the Oracle (85.30).
\textit{Utility preservation:}
As anticipated from Section~\ref{sec:related}, GA-based methods (GA~\cite{yao2024large}, NPO~\cite{zhang2024negative}, WGA~\cite{wang2025rethinking}) suffer significant utility degradation, with perplexity rising to 10--12 on TinyStories. In contrast, FLAT~\cite{wang2024llm}, ASU~\cite{zade2026attention}, STAMP-PI, and STAMP-LR remain stable at approximately 6--8 perplexity, comparable to the Oracle and substantially better than training-based baselines.

\noindent
\textbf{Runtime efficiency:}
All baselines are trained for two epochs, requiring approximately 12 minutes for harmful knowledge removal, 6 minutes for misinformation removal, and 10 minutes for personal data erasure. Being training-free, STAMP reduces this to 7.13, 4.25, and 6.48 minutes, respectively, while STAMP-LR further improves to 4.25, 2.57, and 4.01 minutes, achieving up to $\sim$3$\times$ speedup over training-based methods.

\subsection{RQ2: RePAIR Framework Effectiveness}

We evaluate the full RePAIR pipeline, where intent detection is performed by $\mathcal{M}_{\textit{watchdog}}$, repair code generation by $\mathcal{M}_{\textit{surgeon}}$, and unlearning execution by $\mathcal{M}_{\textit{patient}}$. Experiments are conducted on the WMDP dataset~\cite{li2024wmdp}. We report five metrics: valid code rate, request detection, request satisfaction, IDK rate, and turnaround time in Table~\ref{tab:decomposition}. All metrics except turnaround time are evaluated using Mistral-7B~\cite{Jiang2023Mistral7}, while the user role is simulated via a separate Mistral-7B API instance. Results compare STAMP and STAMP-LR.

Both variants achieve above 96\% across all metrics. The high valid code rate is driven by Qwen2.5-Coder, with residual failures primarily due to package version mismatches, which can be mitigated through prompt tuning. 

Turnaround times 9.36 and 6.50 minutes exceed the RTE reported in Table~\ref{tab:STAMP-Comparison} due to the additional overhead of multi-model orchestration, including $\mathcal{M}_{\textit{watchdog}}$ and $\mathcal{M}_{\textit{surgeon}}$, alongside the unlearning execution.

\subsection{RQ3: Pipeline in Action}
We present qualitative examples of the RePAIR framework in Table~\ref{tab:qualitative_rq3}, illustrating the end-to-end behavior of interactive machine unlearning. Specifically, these examples demonstrate how $\mathcal{M}_{\textit{watchdog}}$ detects unlearning intent, how the system executes the corresponding repair, and how $\mathcal{M}_{\textit{patient}}$ transitions from producing target knowledge to coherent refusal responses. Notably, the model also provides explicit acknowledgment of the unlearning request, ensuring transparency to the user at each stage.
\begin{table}[h!]
\centering
\caption{Comparison of single-layer (Layer~7) vs.\ all-layer activation redirection on Llama-3-8B.}
\label{tab:layer_ablation}
\small
\renewcommand{\arraystretch}{1.3}
\begin{tabular}{l|cccc}
\Xhline{1.15pt}
\textbf{Setting} & \textbf{F-RL}$\downarrow$ & \textbf{R-RL}$\uparrow$ & \textbf{Utility}$\downarrow$ & \textbf{RTE (s)} \\
\Xhline{1.15pt}
Layer~7 only     & 0.00 & 0.85 & 6.07 & 4.36 \\
\rowcolor{gray!15}
All layers       & 0.00 & 0.88 & 6.02 & 15.40 \\
\Xhline{1.15pt}
\end{tabular}
\end{table}
\subsection{Ablation}
\begin{figure}[h!]
    \centering
    \includegraphics[width=0.55\linewidth]{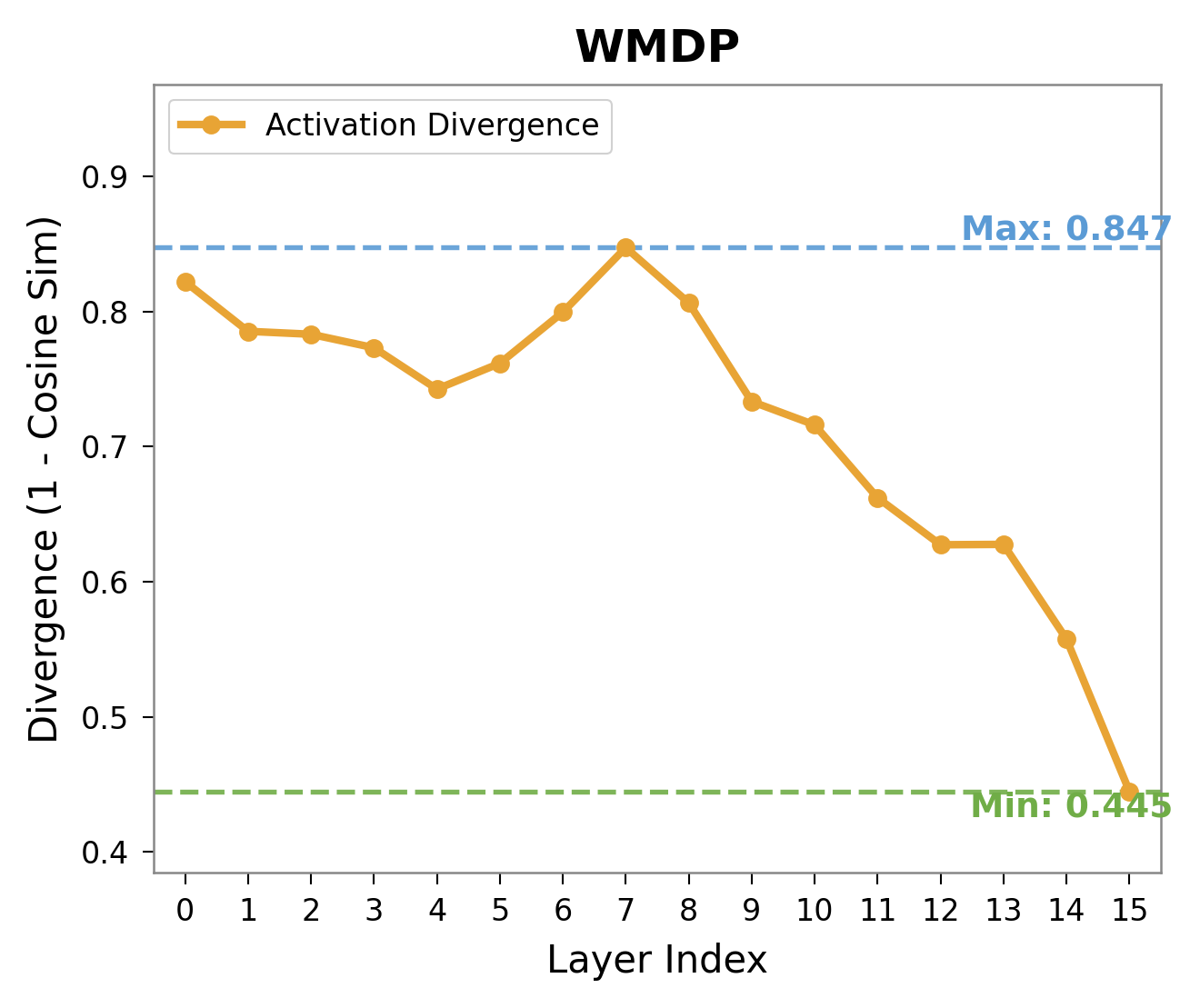}
    \caption{Cosine divergence between WMDP and refusal activations across layers of Llama-3-8B. Layer~7 achieves maximum separation (0.867), motivating its selection as the intervention point.}
    \Description{Line plot showing cosine divergence across layers, with Layer 7 having the highest value, indicating strongest separation between forget and refusal activations.}
    \label{fig:layer_divergence}
\end{figure}

\noindent
\textbf{Layer-wise separation analysis:}
\label{sec:layer_analysis}
We measure cosine divergence between WMDP~\cite{li2024wmdp} and refusal MLP activations across layers of Llama-3-8B, as shown in Figure~\ref{fig:layer_divergence}. Layer~7 achieves the highest separation (0.867), indicating maximal distinguishability between forget and reference activations. Table~\ref{tab:layer_ablation} confirms that intervening at Layer~7 alone matches all-layer redirection in both forgetting and retention, while achieving a $\sim$3.8$\times$ speedup (91s vs.\ 347s).

\noindent
\textbf{Rank analysis:}
STAMP-LR decomposes $\mathbf{X} \approx \mathbf{A}\mathbf{B}$ with rank $r$ (Section~\ref{sec:stamp}). Table~\ref{tab:rank_analysis} varies $r$ on Llama-3-8B. STAMP-LR remains stable and effective for $r \geq 64$. Below this threshold, unlearning becomes incomplete, with residual forget scores, as the low-rank approximation lacks sufficient capacity to capture the full activation structure. All experiments are conducted on the personal data erasure task.

\begin{table}[h!]
\centering
\caption{Effect of rank $r$ on STAMP-LR performance on Llama-3-8B.}
\label{tab:rank_analysis}
\small
\renewcommand{\arraystretch}{1.3}
\begin{tabular}{c|cccc}
\Xhline{1.15pt}
\textbf{Rank ($r$)} & \textbf{F-RL}$\downarrow$ & \textbf{R-RL}$\uparrow$ & \textbf{Utility}$\downarrow$ & \textbf{RTE (mins)} \\
\Xhline{1.15pt}
\rowcolor{gray!15}
8   & 0.00 & 0.72 & 6.83 & 3.15 \\
16  & 0.00 & 0.78 & 6.45 & 3.00 \\
\rowcolor{gray!15}
32  & 0.00 & 0.82 & 6.21 & 3.54 \\
64  & 0.00 & 0.85 & 6.10 & 4.01 \\
\rowcolor{gray!15}
128 & 0.00 & 0.88 & 6.07 & 5.24 \\
\Xhline{1.15pt}
\end{tabular}
\end{table}

\noindent
\textbf{Retain ratio:}
For edge deployment, storing a large retain set is impractical under GDPR and CCPA constraints. We vary the retain buffer $\mathcal{D}_r$ to assess STAMP-LR’s sensitivity Table~\ref{tab:retain_ratio}. Performance remains stable even when $\mathcal{D}_r$ is reduced to 10\% of the full retain set $\mathcal{D}_r^{\mathrm{full}}$. All experiments are conducted on the personal data erasure task.

\begin{table}[H]
\centering
\caption{Effect of retain ratio on STAMP-LR performance on Llama-3-8B.}
\label{tab:retain_ratio}
\small
\renewcommand{\arraystretch}{1.3}
\begin{tabularx}{\columnwidth}{c|>{\centering\arraybackslash}X>{\centering\arraybackslash}X>{\centering\arraybackslash}X>{\centering\arraybackslash}X}
\Xhline{1.15pt}
\textbf{Retain Ratio} & \textbf{F-RL}$\downarrow$ & \textbf{R-RL}$\uparrow$ & \textbf{Utility}$\downarrow$ & \textbf{RTE (s)} \\
\Xhline{1.15pt}
0.10 & 0.00 & 0.88 & 8.83 & 3.12 \\
\rowcolor{gray!15}
0.25 & 0.00 & 0.89 & 8.21 & 3.38 \\
0.50 & 0.00 & 0.87 & 8.74 & 3.61 \\
\rowcolor{gray!15}
0.75 & 0.00 & 0.90 & 7.32 & 3.85 \\
1.00 & 0.00 & 0.90 & 7.07 & 4.01 \\
\Xhline{1.15pt}
\end{tabularx}
\end{table}

\noindent
\textbf{Single-sample unlearning analysis:}
A core requirement of IMU is single-sample forgetting. Table~\ref{tab:single_sample} evaluates all methods under $|\mathcal{D}_f| = 1$ on the harmful knowledge removal task using Llama-3-8B. All baselines are trained for one epoch.

\begin{table}[h!]
\centering
\caption{Single-sample unlearning comparison on Llama-3-8B for harmful knowledge removal ($|\mathcal{D}_f| = 1$).}
\small
\label{tab:single_sample}
\renewcommand{\arraystretch}{1.3}
\begin{tabular}{l|ccc}
\Xhline{1.15pt}
\textbf{Method} & $Acc_f\!\downarrow$ & $Acc_r\!\uparrow$ & \textbf{Utility}$\downarrow$ \\
\Xhline{1.15pt}
GA~\cite{yao2024large}        & 100 & 42.17 & 6.02  \\
\rowcolor{gray!15}
NPO~\cite{zhang2024negative}  & 100 & 48.93 & 5.45  \\
RMU~\cite{li2024wmdp}         & 100 & 39.27 & 6.25  \\
\rowcolor{gray!15}
FLAT~\cite{wang2024llm}       & 100 & 51.63 & 6.05  \\
WGA~\cite{wang2025rethinking} & 100 & 44.51 & 6.13  \\
\rowcolor{gray!15}
ASU~\cite{zade2026attention}  & 100 & 53.12 & 5.48 \\
\midrule\midrule
STAMP                         & 0.00 & 70.13 & 6.55  \\
STAMP-LR                      & 0.00 & 73.27 & 7.00  \\
\Xhline{1.15pt}
\end{tabular}
\end{table}

Training-based baselines fail entirely, i.e., $Acc_f$ remains at 100, as the single-sample gradient signal is overwhelmed by the retain set. In contrast, STAMP and STAMP-LR achieve $Acc_f = 0.00$ with $Acc_r > 70$, confirming their effectiveness for single-sample IMU.
\section{Limitations and Future Work}
This work introduces Interactive Machine Unlearning (IMU) and proposes the RePAIR framework built on STAMP, a training-free, single-sample unlearning method with two variants (STAMP and STAMP-LR). Despite its effectiveness, a few limitations remain.

\noindent
\textbf{Retain data at inference time:} Although STAMP operates with as little as a 10\% retain ratio (Table~\ref{tab:retain_ratio}), it still requires a small replay buffer $\mathcal{D}_r$ to preserve retained knowledge. Storing this buffer on edge devices at inference time is non-trivial and may violate GDPR and CCPA constraints. Methods such as FLAT~\cite{wang2024llm} operate using only $\mathcal{D}_f$ without any replay buffer, suggesting a promising direction toward fully retain-free unlearning, which we leave as future work.

\textbf{Resource constraints at test time:} As shown in Table~\ref{tab:cost_comparison}, training-based methods require significantly more GPU memory than is typically available at inference. While STAMP-LR substantially reduces computational cost, extending the framework to multimodal settings remains an important direction for future work.

\section{Conclusion}
We introduced Interactive Machine Unlearning (IMU), a novel problem setting that enables end users to instruct LLMs to forget targeted knowledge through natural language prompts during inference eliminating the dependency on model service providers. To solve IMU, we proposed RePAIR, a multimodel framework in which $\mathcal{M}_{\textit{watchdog}}$ detects unlearning intent from conversation history, $\mathcal{M}_{\textit{surgeon}}$ generates executable repair code, and $\mathcal{M}_{\textit{patient}}$ undergoes autonomous weight modification. At its core, we introduced the STAMP of training free, single sample unlearning methods such as STAMP and its low-rank variant STAMP-LR which redirect MLP activations toward a refusal subspace via closed form pseudoinverse updates. RePAIR framework is validated accross three unlearning tasks harmful knowledge suppression, misinformation correction, and personal data erasure and achives $\sim$3$\times$ speedup over SoTA methods.


\bibliographystyle{ACM-Reference-Format}
\bibliography{main}

\end{document}